\documentclass[conference]{IEEEtran}
\IEEEoverridecommandlockouts

\usepackage{cite}
\usepackage{amsmath,amssymb,amsfonts}
\usepackage{graphicx}
\usepackage{textcomp}
\usepackage[table,dvipsnames]{xcolor}
\usepackage{graphics}
\usepackage{graphicx}
\usepackage{threeparttable}
\usepackage{multirow} 
\usepackage{hhline}
\usepackage[super]{nth}

\usepackage{algorithm}
\usepackage{algpseudocode}

\usepackage{acronym}
\usepackage{graphicx}
\usepackage{textcomp}
\usepackage{fancyhdr}

\usepackage{siunitx}  

\usepackage{url}

\def\BibTeX{{\rm B\kern-.05em{\sc i\kern-.025em b}\kern-.08em
    T\kern-.1667em\lower.7ex\hbox{E}\kern-.125emX}}

\usepackage{enumitem}    

\newif\ifoutline
\outlinetrue

\renewcommand{\IEEEbibitemsep}{0pt plus 0.5pt}
\makeatletter
\IEEEtriggercmd{\reset@font\normalfont\fontsize{7.9pt}{8.60pt}\selectfont}
\newcommand*\titleheader[1]{\gdef\@titleheader{#1}}
\AtBeginDocument{%
  \let\st@red@title\@title
  \def\@title{%
    \bgroup\normalfont\normalsize\centering\@titleheader\par\egroup
    \vskip1ex\st@red@title}
}
\makeatother
\IEEEtriggeratref{1}

\titleheader{\vspace{-30pt}To appear at the 27th Design, Automation and Test in Europe Conference (DATE'24), Mar 25-27 2024, Valencia, Spain.}
\title{On-sensor Printed Machine Learning Classification \\ via Bespoke ADC and Decision Tree Co-Design}

\begin{document}

\bstctlcite{IEEEexample:BSTcontrol} 
\setlength{\abovedisplayskip}{2ex}
\setlength{\belowdisplayskip}{2ex}

\author{
    \IEEEauthorblockN{
        Giorgos Armeniakos\IEEEauthorrefmark{1},
        Paula L. Duarte\IEEEauthorrefmark{2},
        Priyanjana Pal\IEEEauthorrefmark{2},
        Georgios Zervakis\IEEEauthorrefmark{3},\\
        Mehdi B. Tahoori\IEEEauthorrefmark{2},
        Dimitrios Soudris\IEEEauthorrefmark{1}
    }
    \IEEEauthorblockA{
        \IEEEauthorrefmark{1}National Technical University of Athens, GR,
        \IEEEauthorrefmark{3}University of Patras, GR,
        \IEEEauthorrefmark{2}Karlsruhe Institute of Technology, DE
    }
    \IEEEauthorblockA{
        \IEEEauthorrefmark{1}\{armeniakos, dsoudris\}@microlab.ntua.gr,
        \IEEEauthorrefmark{2}\{paula.duarte, priyanjana.pal, mehdi.tahoori\}@kit.edu,
        \IEEEauthorrefmark{3}zervakis@upatras.gr
    }
}

\maketitle

\begin{abstract}
Printed electronics (PE) technology provides cost-effective hardware with unmet customization, due to their low non-recurring engineering and fabrication costs. 
PE exhibit features such as flexibility, stretchability, porosity, and conformality, which make them a prominent candidate for enabling ubiquitous computing.
Still, the large feature sizes in PE limit the realization of complex printed circuits, such as machine learning classifiers, especially when processing sensor inputs is necessary, mainly due to the costly analog-to-digital converters (ADCs).
To this end, we propose the design of fully customized ADCs and present, for the first time, a co-design framework for generating bespoke Decision Tree classifiers.
Our comprehensive evaluation shows that our co-design enables self-powered operation of on-sensor printed classifiers in all benchmark cases.
\end{abstract}


\begin{IEEEkeywords}
Co-design, Decision Tree, Low-Power, Machine Learning, Printed Electronics
\end{IEEEkeywords}

\section{Introduction}



Printed electronics technology, capturing substantial attention, holds promise for computing in various domains yet to embrace it. Examples include smart packaging, disposables (e.g., packaged foods), FMCG, in-situ monitoring, and healthcare products like smart bandages~\cite{Bleier:ISCA:2020:printedmicro}. These domains impose rigorous prerequisites for ultra-low-cost fabrication, unattainable by silicon-based systems. Furthermore, aside from their high costs, silicon systems lack the stretchability, porosity, flexibility, and conformality inherent in printed electronics.

While printed circuits exhibit the potential to meet the demands of ultra-low cost (even sub-cent levels) due to their inexpensive additive manufacturing processes,
the relatively large feature sizes, inherent in printed electronics, pose a challenge in realizing large-scale integration circuits for on-sensor processing.
This becomes particularly relevant for Machine Learning (ML) classification circuits, as classification is the primary task for numerous applications in aforementioned domains~\cite{Mubarik:MICRO:2020:printedml,arm2023codesign}. 
Such applications have to classify (analog) data collected from printed sensors to extract useful information.

Significant recent research has focused on addressing the intrinsic constraints of printed electronics, aiming to enable the realization of printed-battery-powered printed ML classifiers~\cite{Mubarik:MICRO:2020:printedml,arm2023codesign,arm2023crossapprox,Kokkinis:DATE2023,Weller:2021:printed_stoch,isqed_dt}. 
In~\cite{Mubarik:MICRO:2020:printedml}, the authors leverage the substantial customization capabilities stemming from the low fabrication costs of printed circuits, and explored the concept of \emph{bespoke} ML circuits that are tailored to a specific model.
The term ``bespoke'' signifies fully customized circuit implementations, adapting to individual ML models and dataset, and can be obtained by hardwiring all model parameters in the circuit design.
This level of customization is infeasible in conventional silicon-based systems~\cite{Mubarik:MICRO:2020:printedml}.
While bespoke design indeed yields significant area and power gains, further improvements are mandatory to enable printed-battery-powered ML classifiers.
The works in~\cite{arm2023codesign,arm2023crossapprox,Kokkinis:DATE2023} 
and~\cite{Weller:2021:printed_stoch} combined bespoke design with approximate~\cite{Shafique:DAC2016} and stochastic~\cite{stochastic} computing, respectively, targeting Multilayer Perceptrons and/or Support Vector Machines.
However, in most cases, the obtained gains are either inadequate or come at the cost of a considerable accuracy degradation.
Leveraging the fact that in printed applications the classification tasks and datasets are relatively simple~\cite{Mubarik:MICRO:2020:printedml}, the authors in~\cite{Mubarik:MICRO:2020:printedml,isqed_dt} utilize Decision Trees due to their implementation simplicity.
Combining Decision Trees with bespoke design and approximate computing,~\cite{isqed_dt} delivered promising results.
Despite the impressive results demonstrated in the aforementioned works, they all overlook the substantial--potentially dominant--power and area overheads associated with the mandatory analog-to-digital converters (ADCs) for processing sensor data.
Thus, their efficiency remains unclear.


In this work, we address the limitation of the current state-of-the-art.
Our focus goes beyond printed-battery operation; instead, we target the design of self-powered (i.e., using printed energy harvesters) printed classifiers, accounting also for the cost of the ADCs.
To achieve this, we adopt Decision Trees as our classification algorithm and propose a model-circuit co-design framework.
To maximize hardware efficiency, we implement bespoke Decision Tree classifiers based on the unary numeric system~\cite{ugemm:isca20}.
As we explain later, this approach may reduce a Decision Tree to a simple two-stage logic.
Furthermore, we leverage the high customization in printed circuits and propose the design of bespoke ADCs tailored to our Decision Tree architecture by retaining only the bare minimum of ADC comparators.
Finally, we propose and implement an ADC-aware Decision Tree training that identifies parameters that minimize the ADCs cost while still adhering to accuracy requirements.
Our evaluation demonstrates that, compared to the state-of-the-art baseline, we reduce the area and power of printed Decision Trees by on average $8.6$x and $12.2$x, respectively, for less than $1$\% accuracy degradation.
\textbf{Our novel contributions in this work are as follows:}
\begin{enumerate}[topsep=0pt,leftmargin=*]
    \item This is the first work that considers and investigates the impact of ADCs in designing digital printed ML classifiers.
    \item We propose the first ADC-aware co-design framework dedicated to bespoke printed Decision Trees\footnote{Available at \url{https://github.com/garmeniakos/Ax-Printed-ML-Classifiers}.}.
    \item Our framework enables for the first time self-powered on-sensor digital ML classifiers in printed electronics.
    
\end{enumerate}

\section{Background}
\subsection{Printed Electronics}
Much akin to color printing, Printed Electronics (PE) technology follows an additive manufacturing process for electronic components, utilizing diverse printing techniques like roll-to-roll and jet printing~\cite{book}.
Tailored to the printing materials and substrates used, printed circuits present multiple benefits such as non-toxicity, lightness, and flexibility.
Also, due to the additive nature of PE, 
the production of printed circuits can be achieved at an exceptionally low cost.
This advantage greatly simplifies the production of innovative products by enabling cost-effective manufacturing of customized, purpose-specific electronic devices in small quantities.
Though, due to the low precision printing, PE exhibits low integration density (orders of magnitude lower than silicon VLSI), large feature sizes, and elevated device latencies.
Nevertheless, in target domains, the performance and computation requirements are generally quite modest, e.g., sampling rate of a few Hz and few bits precision~\cite{Henkel:ICCAD2022:expedition}. 
These requirements could be effectively met by printing technologies, while still adhering to acceptable area and energy constraints.
In this work, we consider the inorganic Electrolyte-Gated FET (EGFET) technology~\cite{Bleier:ISCA:2020:printedmicro} that features below $1$V supply voltage and high mobility characteristics, unlike organic counterparts, making it a suitable match for low-power and energy-harvested applications.

\subsection{Flash ADC}
ADCs play a crucial role in modern electronics by converting analog signals, such as sensor data, into digital for further processing.
We focus on Flash ADC because of its implementation simplicity, especially for very low precision requirements as in printed applications, and its minimal latency.
Flash ADCs excel at delivering rapid data conversion, ensuring that data is promptly available even during short energy availability windows.
Additionally, despite consuming high power during conversion, they mainly operate in a low-power state, thereby minimizing average power and energy consumption, aligning well with energy-harvesting approaches.



\begin{figure}
    \centering
    \includegraphics[width=\columnwidth]{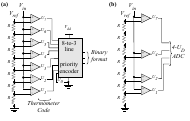}\vspace{-3ex}
    \caption{Schematic of: a) conventional 3-bit Flash ADC and b) an example of an equivalent bespoke ADC with four unary digits of output.}
    \label{fig:adc}\vspace{-2ex}
\end{figure}

A typical 3-bit flash ADC block diagram is shown in Fig.~\ref{fig:adc}a.
A key component of a Flash ADC is a bank of comparators.
The number of comparators in the array corresponds to the desired resolution of the ADC.
For an N-bit ADC, $2^N\!-\!1$ comparators are required.
The input analog voltage (Vin) is connected to the non-inverting terminals of all the comparators. 
The reference voltage range is divided into $2^N$ segments to cover the entire range of the input analog voltage.
Each comparator is connected to a reference voltage (Vref) that corresponds to the midpoint of each segment.
When an analog input voltage Vin is applied to the ADC, each comparator in the array compares Vin with its corresponding Vref.
If Vin is larger than Vref, the comparator outputs a high signal (1); otherwise, it outputs a low signal (0).
The comparators outputs form the digital representation of Vin in a thermometer code, which is then processed by a fast priority encoder to produce the corresponding binary value.

\subsection{Unary Decision Tree Circuits}
Unary computing traditionally operates on serial bitstreams using extremely simple logic~\cite{ugemm:isca20}.
State-of-the-art unary implementations typically rely on rate or temporal logic~\cite{ugemm:isca20} and handle data in the form of serial bitstreams~\cite{racetree:asplos19}.
While these prior works present efficient unary computation units, 
they require expensive converters~\cite{bazargan:unary:tvlsi18}
and mandate multi-cycle operation, which is prohibitive in PE~\cite{Mubarik:MICRO:2020:printedml}.
On the other hand, we exploit the unmatched customization capabilities of PE and design fully parallel, unary-based Decision Tree classifiers.

\section{Our Proposed Co-design Framework}\label{sec:codesign}

This section describes our co-design framework for generating printed Decision Trees.
In brief, we first introduce the architecture of a fully-parallel Decision Tree based on the unary representation and highlight its area and power benefits over non-unary approaches.
Then, we analyze our flash ADC design tailored for such architectures, and we describe our 
ADC-aware Decision Tree training that enables minimizing the ADCs hardware cost while maintaining high accuracy.

\subsection{Parallel Unary Decision Trees}\label{sec:unaryDT}
In unary coding (or else thermometer code), an $N$-bit binary number is represented by a code of length $2^N\!-\!1$. 
The count of '1's in the unary code corresponds to the value being represented.
Unary coding can express many types of numbers, such as integers, fixed-point, etc.
For instance:
\begin{align}
    0011111_\text{U} &= 101_{2} = 5_{10} \\   
    .111_\text{U} &= .11_{2} = .75_{10} \notag
\end{align}

A parallel unary format is generally not preferred due to its increased size to represent a number ($N$ vs $2^N\!-\!1$ bits)~\cite{hybrid_unary_bazargan:2023}. 
However, as we show hereafter, this does not hold in our case and thus, we investigate and propose the implementation of parallel unary printed Decision Trees.
As aforementioned, the unmatched customization in printed circuits enables hardcoding the model's parameters. 
As a result, a comparison $I\geq C$ where $I$ is an input and $C$ is a model parameter, becomes $I\geq .1011_2$ assuming that $.1011_2$ is the trained value of $C$.
In unary format $C$ can be written as $.000011111111111_\text{U}$.
Therefore, although parallel unary representation doesn't typically make sense in conventional architectures due to the exponentially increased number of bits to be compared, in bespoke Decision Trees
the comparator is essentially reduced to simply checking a bit from the input:
\begin{equation}
    I\geq .1011_2  \xrightarrow{\text{Unary}}  I \geq .000011111111111_\text{U} \equiv I[11]
    \label{eq:comp}
\end{equation}
In other words, if the \nth{11} bit of I is `$1$' then the initial inequality $I\geq C$ is true.
In general, if the most significant `$1$' of $C$ is at bit position $k$, then  $I\geq C \equiv I[k]$.
In addition, the inequality is directly computed ($I$ is in parallel format), eliminating the need to wait for bit $k$ to arrive from a serial input.
Similar relations are derived for all comparisons: $I > C \equiv I[k+1]$, $I < C \equiv !(I[k])$, and $I \leq C \equiv !(I[k+1])$.
Note that in unary format if $I[k]=1$, then $I[j]=1$ $\forall j\leq k$.

The analysis above highlights that in a bespoke Decision Tree, if the inputs are provided in a parallel unary format, all Tree comparators can be removed. 
This is perfectly in line with the limited hardware resources of printed circuits. 
It is crucial to note that a serial (e.g., temporal) unary representation would not only enforce a printed-unfriendly multi-cycle operation but would also introduce a significant hardware overhead such as control circuitry and potentially numerous registers.

\begin{figure}
    \centering
    \includegraphics[width=\columnwidth]{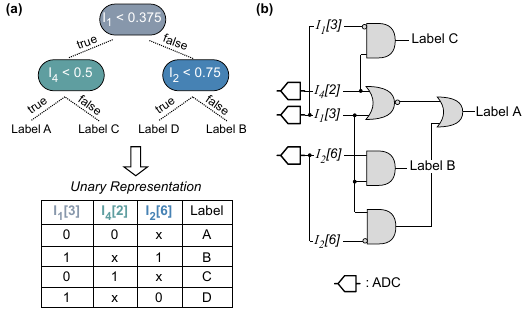}
    \caption{Illustration of how a conventional Decision Tree (a) is translated into unary format, represented by a set of unary digits. This example assumes Q0.4 formatted values, e.g., $0.75\rightarrow6$. (b) depicts the simplified schematic.}
    \label{fig:unarycomp}\vspace{-2ex}
\end{figure}

Fig.~\ref{fig:unarycomp} presents an example architecture of a Decision Tree when its inputs are available in a parallel unary representation.
Instead of traditional comparisons, bespoke unary DTs can now be viewed as a simple logic over a set of unary digits corresponding to the trained parameters (e.g., $I_1$[3], $I_4$[2], $I_2$[6]).
The truth table for each label of the unary Decision Tree example is also depicted.
As shown, only a few gates are sufficient.
Each label is obtained through a simple two-level logic (e.g., \texttt{AND}-\texttt{OR}) and each input signal in this two-level logic denotes a node in the Decision Tree.

\subsection{Bespoke ADCs}\label{subsec:adc}

As shown in the previous section, ensuring the availability of inputs in a parallel unary representation minimizes the hardware overheads of printed Decision Tree classifiers.
As shown in Fig.~\ref{fig:adc}a, the flash ADC calculates an intermediate result, which is the thermometer code of the input signal (each $U_i$ denotes if Vin is larger than Vref).
As a result, by simply removing the encoder, we not only decrease the ADC's hardware requirements but also achieve our objective: transforming the sensor input into a parallel unary representation before supplying it to the Decision Tree classifier.
This also further justifies our flash ADC consideration.

To further improve the hardware-efficiency of our ADCs, we also implement them in a fully customized manner.
The simplified Decision Tree design described above (e.g.,  Fig.~\ref{fig:unarycomp}) relies on a parallel unary representation of its inputs.
Though, as indicated by~\eqref{eq:comp}, each comparison requires only a specific input bit.
Consequently, the remaining unary digits, if not needed for another comparison, can be discarded and don't have to be generated.
An illustrative example of a bespoke ADC is presented in Fig.~\ref{fig:adc}b.
In this example, we assume that the respective input is involved in multiple comparisons and is compared against four different parameters.
Hence, only four unary digits need to be calculated.
Without any loss of generality, this example assumes that the \nth{1}, \nth{2}, \nth{4}, and \nth{7} unary digits are required for the comparisons.
Fig.~\ref{fig:adc}b presents the architecture of the corresponding ``4-$U_D$'' (four output unary digits) ADC.
To generate the specific ADC, we only need to retain the corresponding four comparators, and we can eliminate the remaining three comparators and the encoder.
In general, bespoke ADC design entails retaining only the resistors and the bare-minimum comparators required.

\begin{figure}
    \centering
    \includegraphics[width=\columnwidth]{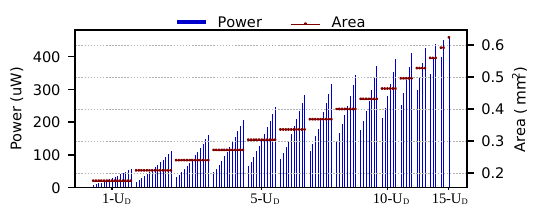}
    \caption{The Area and power of (4-bit) bespoke ADCs w.r.t. their output unary digits. The first values correspond to 1-output ADCs, while the last one to 15-output ADC. Different points denote different output digits. Selected output digits are in sequential order, i.e., ``$U_1$-$U_2$'' 2-$U_D$ ADC is followed by ``$U_2$-$U_3$'' 2-$U_D$ ADC and so on, only to showcase the behavior of power.}
    \label{fig:powerADC}\vspace{-3ex}
\end{figure}

The number of comparators, as well as the specific ones to be retained, is determined by the trained Decision Tree parameters.
Considering a conventional 4-bit ADC, Fig.~\ref{fig:powerADC} presents how the area and power characteristics of our bespoke ADCs scale w.r.t. the number and position of their output unary digits.
In Fig.~\ref{fig:powerADC}, the number of output digits ranges from $1$ to $15$.
For example, a 2-$U_D$ ADC denotes a bespoke ADC with only two outputs, while a 15-$U_D$ ADC indicates the retention of all comparators.
To showcase the power behavior, a few representative examples are presented w.r.t. the specific outputs that are retained.
The area and power of the conventional 4-bit ADC are $11$mm$^2$ and $0.83$mW, respectively.
To obtain the area and power values, we designed the ADCs in Cadence Virtuoso using the EGFET Process Design Kit (PDK)~\cite{Bleier:ISCA:2020:printedmicro} and conducted SPICE simulations.
The simulations are conducted with a voltage supply of $1$V.
As shown in Fig.~\ref{fig:powerADC}, the area of a bespoke ADC solely depends on the number of output unary digits of the ADC.
Specifically, the area scales linearly along with the number of the comparators.
On the other hand, power consumption also depends on which outputs are selected.
For example, in a 4-$U_D$ bespoke ADC, power consumption ranges from $47$uW up to $205$uW, i.e., $4.4\times$ increase.
It is observed that the power is substantially decreased when lower-order outputs are selected.
This can be attributed to the lower Vref of the lower-order comparators.
Fig.~\ref{fig:powerADC} shows  a linear increase in the comparators power consumption as we move towards higher-order outputs.
The key takeaways from this analysis are twofold.
First, bespoke ADCs provide significant hardware improvements over conventional ADCs.
Second, the area and power consumption of a bespoke ADC are highly influenced by its outputs.
By minimizing the number of outputs for each ADC, we can minimize its area, and by carefully selecting the specific outputs, we can further boost its power efficiency.

\subsection{ADC-aware Decision Tree Training}


As illustrated in Fig.~\ref{fig:unarycomp}b, assuming parallel unary inputs, the logic of the Decision Tree is inherently simplified. 
Hence, since the hardware-efficiency of this two-level logic is mostly determined by the Decision Tree depth hyperparameter, our training methodology focuses on optimizing the cost of the ADCs.
As demonstrated in Section~\ref{subsec:adc}, the hardware efficiency of the ADCs, and consequently of the entire classifier, is determined by the trained parameters of the Decision Tree (i.e., output unary digits of each ADC).
Hence, carefully selecting the Decision Tree parameters in an ADC-aware manner, while still achieving high classification accuracy, is essential for enabling power-autonomous printed Decision Trees.

\begin{algorithm}[t!]
\caption{ADC-aware Decision Tree Training Pseudocode}\label{alg:train}
\footnotesize
\textbf{Input:} 1) Dataset, 2) Tree Depth, 3) Gini Threshold $\tau$ \\
\textbf{Output:} Trained Decision Tree and Classification accuracy
\begin{algorithmic}[1]
\State $\mathcal{DT} = \varnothing$ \#\textit{selected split nodes} 
\State \textbf{for} $0 \leq$ node $<$ Total nodes \textbf{do}
\State \hspace{3mm} \textbf{for} $\forall I_i \in$ Input Features \textbf{and} $\forall C$ value in dataset for $I_i$ \textbf{do}
\State \hspace{6mm} calculate $\mathrm{Gini}(I_i,C)$
\State \hspace{3mm} $G$ = minimum Gini score
\State \hspace{3mm} $\mathcal{S} = \{(I_i,C) \;|\; \mathrm{Gini}(I_i,C)\leq G+\tau, \forall (I_i,C)\}$
\State \hspace{3mm} $\mathcal{S_Z} =  \{(I_i,C) \in \mathcal{S} \;|\; (I_i,C) \in \mathcal{DT}\}$
\State \hspace{3mm} $\mathcal{S_M} =  \{(I_i,C) \in \mathcal{S} \;|\; \exists (I_i,C^\prime) \in \mathcal{DT},\, C\,\neq\,C^\prime\}$
\State \hspace{3mm} $\mathcal{S_H} =  \{(I_i,C) \in \mathcal{S} \;|\; \exists! (I_i,C^\prime) \in \mathcal{DT},\, \forall C^\prime\}$
\State \hspace{3mm} \textbf{if} $\mathcal{S_Z}\,\neq\, \varnothing$ \textbf{then}
\State \hspace{6mm} $g_m$ = calculate minimum Gini score $\forall (I_i,C) \in \mathcal{S_Z}$
\State \hspace{6mm} $split=\mathrm{random}(\{(I_i,C) \in \mathcal{S_Z} \;|\; \mathrm{Gini}(I_i,C)=g_m\})$
\State \hspace{3mm} \textbf{else}
\State \hspace{6mm} \textbf{if} $\mathcal{S_M}\,\neq\, \varnothing$ \textbf{then} $\mathcal{Z} = \mathcal{S_M}$ \textbf{else} $\mathcal{Z} = \mathcal{S_H}$
\State \hspace{6mm} $c_m = \mathrm{min}(\{C \;|\; \forall (I_i,C) \in \mathcal{Z}\})$
\State \hspace{6mm} $\mathcal{U} =  \{(I_i,C) \in \mathcal{Z} \;|\; C=c_m\}$
\State \hspace{6mm} $g_m$ = calculate minimum Gini score $\forall (I_i,C) \in \mathcal{U}$
\State \hspace{6mm} $split=\mathrm{random}(\{(I_i,C) \in \mathcal{U} \;|\; \mathrm{Gini}(I_i,C)=g_m\})$
\State \hspace{3mm} $\mathcal{DT} = \mathcal{DT} \cup \{split\}$
\end{algorithmic}
\end{algorithm}


To achieve this, we propose and implement an ADC-aware Decision Tree training approach.
Our primary objective is to minimize the number of comparators induced by the ADCs.
This is accomplished by minimizing the number of unique inputs involved among the total comparisons (i.e., minimizing the number of ADCs) and, for each remaining input, by minimizing the number of different parameters, it is compared against, in all the comparisons involved in.
Our secondary objective is to select more power-efficient values/parameters for each comparison (i.e., optimize the order of the output digits in the ADC).
Since feasibility is the foremost requirement for printed ML circuits, prioritizing it over strict accuracy constraints is a typical procedure~\cite{arm2023codesign}.
Consequently, we also explore the trade-off between some accuracy degradation and the potential for additional hardware gains.

Our ADC-aware training essentially trains a Decision Tree using the \emph{Gini index}~\cite{gini} cost function.
Gini is used to evaluate a split in the dataset.
The latter involves one input feature (e.g., $I_i$ in Fig.~\ref{fig:unarycomp}a) and a trainable parameter $C$ that will be compared with (i.e., output unary digit of the ADC of $I_i$).
We modify typical Gini-based Decision Tree training to incorporate ADC-awareness as follows.
Initially, our algorithm seeks a split node and evaluates the Gini score for all possible combinations between input features and their corresponding values in the training dataset.
At this point, ADC-unaware training would randomly select one combination among those with the best (minimum) Gini score.
Assuming $G$ is the best Gini score computed, we form a set of candidate split pairs $\mathcal{S}=\{(I_i,C)| \mathrm{Gini}(I_i,C)\leq G+\tau \}$, with $\tau$ being a training hyperparameter.
Next, we group the pairs $(I_i, C) \in \mathcal{S}$ into three sets based on the hardware induced from their selection.
\begin{enumerate}[topsep=0pt,leftmargin=*]
\item  $\mathcal{S}_Z$ (zero-cost): if $(I_i, C)$ has been previously selected at a split node, it won't require additional hardware, only additional wiring.
\item  $\mathcal{S}_M$ (medium-cost): if $(I_i, C^\prime)$ with $C^\prime\neq C $ has been previously selected at a split node, then the same ADC for $I_i$ is reused, but a new comparator is added to that ADC because a different output digit is required.
Since the area of our bespoke ADCs is linear with the number of comparators, all these pairs induce the same overhead, as each of them will add one comparator to one ADC.
\item  $\mathcal{S}_H$ (high-cost): if no pair $(I_i, C^\prime)$ has been previously selected at a split node, i.e., $I_i$ is selected for the first time.
These pairs induce the highest (and same) area overhead because a new ADC with one comparator is required.
\end{enumerate}
Among these sets, we select the first non-empty one based on the order listed above.
If $\mathcal{S}_M$ or $\mathcal{S}_H$ is chosen, we then identify the $(I_i, C)$ pair in that set that results in the lowest power overhead.
This can be accomplished by selecting the $(I_i, C)$ pair with the minimum $C$ value since it will necessitate the lowest-order output digit and, consequently, the induced comparator will have the lowest power consumption (see Section~\ref{subsec:adc}).
If multiple pairs feature the same minimum $C$ value, or if $\mathcal{S}_Z$ is chosen, we select the pair with the best Gini score, or random one if the Gini scores are equal.
Our ADC-aware training assumes user-defined fixed depth and $\tau$ hyperparameters.
Our algorithm identifies the most hardware-efficient split at each node. While our approach is inherently greedy, it introduces hardware-awareness into the traditionally employed, also greedy, Gini-based training. 
Still, alternative heuristic approaches could also be used.
Algorithm~\ref{alg:train} provides an abstract overview of our proposed training methodology.

The hyperparameter $\tau$ (if higher than $0$) may lead to some accuracy degradation, as a pair with a Gini score higher than the best score might be selected.
However, $\tau$ increases the size of the set $\mathcal{S}$, thereby increasing the chances of finding a more hardware-efficient $(I_i, C)$.
$\tau=0$ will not affect the accuracy.


\section{Results and Analysis}\label{sec:eval}

\begin{table}[t]
\setlength\tabcolsep{3.7pt}
\footnotesize
\centering
\renewcommand{\arraystretch}{1.2}
\caption{Evaluation of the baseline bespoke Decision Trees.
}
\begin{tabular}{l|c|c|c|cr|cr}
\hline
\multicolumn{1}{c|}{\multirow{2}{*}{\textbf{Dataset}}} &
  \multirow{2}{*}{\textbf{\begin{tabular}[c]{@{}c@{}}Acc\\ (\%)\end{tabular}}} &
  \multirow{2}{*}{\textbf{\#Comp.}} &
  \multirow{2}{*}{\textbf{\#Inputs}} &
  \multicolumn{2}{c|}{\textbf{Area (mm$^2$)}} &
  \multicolumn{2}{c}{\textbf{Power (mW)}} \\
\multicolumn{1}{c|}{} &      &     &    & ADCs  & \multicolumn{1}{c|}{Total} & ADCs  & \multicolumn{1}{c}{Total} \\ \hline
\cellcolor[HTML]{F0F0F0}Whitewine             & \cellcolor[HTML]{F0F0F0}52.8 & \cellcolor[HTML]{F0F0F0}207 & \cellcolor[HTML]{F0F0F0}11 & \cellcolor[HTML]{F0F0F0}17.3 & \cellcolor[HTML]{F0F0F0}261.3                      & \cellcolor[HTML]{F0F0F0}5.4  & \cellcolor[HTML]{F0F0F0}14.6                      \\
Cardio                & 90.6 & 85  & 19 & 22.3 & 114.4                      & 9.1  & 12.5                      \\
\cellcolor[HTML]{F0F0F0}Arrhythmia            & \cellcolor[HTML]{F0F0F0}62.7 & \cellcolor[HTML]{F0F0F0}39  & \cellcolor[HTML]{F0F0F0}21 & \cellcolor[HTML]{F0F0F0}23.5 & \cellcolor[HTML]{F0F0F0}79.9                       & \cellcolor[HTML]{F0F0F0}10.0 & \cellcolor[HTML]{F0F0F0}12.0                      \\
Balance-Scale         & 77.7 & 15  & 4  & 12.9 & 30.6                       & 2.2  & 2.9                       \\
\cellcolor[HTML]{F0F0F0}Vertebral-3C          & \cellcolor[HTML]{F0F0F0}86.0 & \cellcolor[HTML]{F0F0F0}7   & \cellcolor[HTML]{F0F0F0}5  & \cellcolor[HTML]{F0F0F0}13.6 & \cellcolor[HTML]{F0F0F0}16.8                       & \cellcolor[HTML]{F0F0F0}2.5  & \cellcolor[HTML]{F0F0F0}2.8                       \\
Seeds                 & 90.5 & 23  & 5  & 13.6 & 27.3                       & 2.5  & 3.2                       \\
\cellcolor[HTML]{F0F0F0}Vertebral-2C          & \cellcolor[HTML]{F0F0F0}87.1 & \cellcolor[HTML]{F0F0F0}7   & \cellcolor[HTML]{F0F0F0}5  & \cellcolor[HTML]{F0F0F0}13.6 & \cellcolor[HTML]{F0F0F0}16.4                       & \cellcolor[HTML]{F0F0F0}2.5  & \cellcolor[HTML]{F0F0F0}2.8                       \\
Pendigits             & 95.0 & 215 & 16 & 20.4 & 268.7                      & 7.7  & 17.2                      \\ \hline
\end{tabular} \vspace{-3ex}
\label{tab:baselines}
\end{table}


In this section, we evaluate our printed Decision Tree classifiers, first examining hardware gains from our bespoke ADC design and ADC-aware Decision Tree training. 
Evaluation is based on 8 datasets listed in Table~\ref{tab:baselines}.
These datasets are selected  for two primary reasons: i) to facilitate direct comparisons with the state-of-the-art~\cite{Mubarik:MICRO:2020:printedml,isqed_dt}, and ii) because these datasets utilize sensor inputs suitable for printed applications~\cite{Mubarik:MICRO:2020:printedml,Weller:2021:printed_stoch}.
The datasets are obtained from the the UCI ML repository\cite{uci}.
Normalized inputs in the range $[0, 1]$ are used for training/testing with a random $70$\%/$30$\% split.
Synopsys Design Compiler and PrimeTime analyze the digital part of circuits, all operated at $20$Hz, a common frequency aligned with typical performance of target PE applications~\cite{Bleier:ISCA:2020:printedmicro,arm2023codesign}.

As our evaluation baseline, we consider the fully parallel bespoke Decision Trees designed as described in~\cite{Mubarik:MICRO:2020:printedml}.
The minimum tree depth (up to $8$) that achieves the maximum accuracy is used for each model and the input precision is set to $4$ bits, since this is the value delivered close to floating-point accuracy for all datasets.
Table~\ref{tab:baselines} summarizes the accuracy and hardware overheads of the baseline Decision Trees~\cite{Mubarik:MICRO:2020:printedml}.
Specifically, Table~\ref{tab:baselines} reports the total area and total power requirements for each Decision Tree classifier as well as the respective values for the ADCs.
As shown, the average total area and power consumption are $102$mm$^2$ and $8.5$mW, respectively.
Notably, all the classifiers exhibit power demands that exceed the capabilities ($>\!2$mW) of printed energy harvesters~\cite{pintedharvester}.
Consequently, none of these circuits can be self-powered.
Finally, it's worth noting that, on average, for the Decision Trees in Table~\ref{tab:baselines}, $40$\% of the total area and $74$\% of the total power consumption is attributed to the ADCs.

\begin{figure}
    \centering
    \includegraphics[width=\columnwidth]{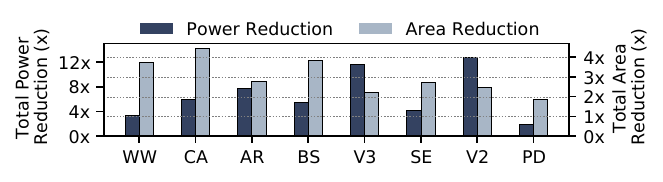}
    \vspace{-5ex}
    \caption{Total area and power reduction (x) compared to the baseline designs~\cite{Mubarik:MICRO:2020:printedml} (i.e., vs Table~\ref{tab:baselines}). For our printed Decision Trees only our proposed bespoke ADCs and parallel unary architecture are considered.}
    \label{fig:adconly}\vspace{-3ex}
\end{figure}
\begin{figure}
    \centering
    \includegraphics[width=\columnwidth]{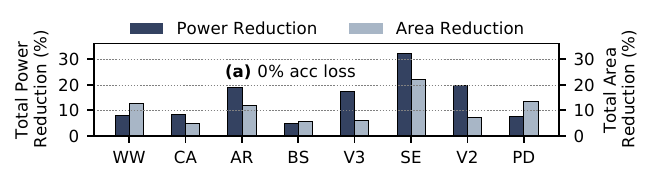}
    \includegraphics[width=\columnwidth]{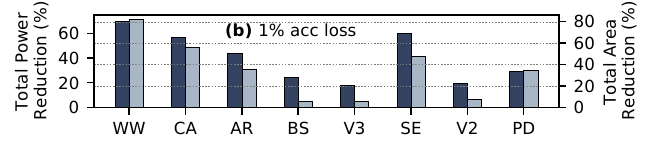} 
    \includegraphics[width=\columnwidth]{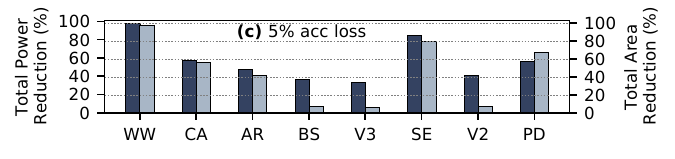} 
    \vspace{-5ex}
    \caption{Evaluation of the additional hardware gains delivered by our ADC-aware training.
    Total area and power reductions (\%) of our printed DTs w/ and w/o our ADC-aware training are compared (i.e., vs Fig.~\ref{fig:adconly}).
    Three accuracy loss constraints are considered: a) 0\% (i.e., no accuracy loss), b)  1\%, c) 5\%.}
    \label{fig:barplots}\vspace{-2ex}
\end{figure}

Fig.~\ref{fig:adconly} depicts the area and power gains achieved by solely considering our bespoke ADCs along with the parallel unary Decision Tree design, i.e., the same ADC-unaware trained model used in~\cite{Mubarik:MICRO:2020:printedml}.
In Fig.~\ref{fig:adconly}, values are reported w.r.t. the baseline~\cite{Mubarik:MICRO:2020:printedml} (see Table~\ref{tab:baselines}).
As shown, given that the overall area and power consumption of the printed Decision trees are predominantly governed by the ADCs, employing our bespoke ADC design delivers substantial hardware gains.
Furthermore, as explained in Section~\ref{sec:codesign}, using our ADCs streamlines the classifier's implementation, resulting in a simplified two-stage logic and additional hardware savings over the baseline~\cite{Mubarik:MICRO:2020:printedml}.
Specifically, compared to~\cite{Mubarik:MICRO:2020:printedml}, the achieved area and power reduction are $3.0$x and $6.6$x on average, respectively. 

Next, in Fig.~\ref{fig:barplots} we investigate the impact of our ADC-aware training.
The area and power gains in Fig.~\ref{fig:barplots} are reported w.r.t. the area and power of the designs of  Fig.~\ref{fig:adconly} (i.e., over the simplified Decision Trees with our bespoke ADCs).
For this analysis, we consider three accuracy loss thresholds: $0$\%, $1$\%, and $5$\%.
Specifically, we conduct a brute-force exploration of hyperparameters, including $\tau$ values ranging from $0$ to $0.03$ in increments of $0.005$ and depth values from $2$ to $8$ with a step of $1$.
The rather simple classification tasks in printed electronics~\cite{Mubarik:MICRO:2020:printedml} and the independence of different trainings (i.e., they can run in parallel), enable this exploration to be rapidly conducted.
The average execution time is only $6$ min on an Intel Xeon 6138 server with $256$GB RAM.
As demonstrated in Fig.~\ref{fig:barplots}a, for \textit{no accuracy loss}, our ADC-aware training leads to an average reduction of $11$\% in area and $15$\% in power when compared to using the conventional ADC-unaware training.
Similarly, for only $5$\% accuracy loss the average area and power reduction increase to $45$\% and $57$\%, respectively.
\begin{table}[t]

\setlength\tabcolsep{4pt}
\footnotesize
\centering
\renewcommand{\arraystretch}{1.2}
\caption{Evaluation of our Decision Trees for up to 1\% Accuracy Loss.}
\begin{threeparttable}
\begin{tabular}{l|cc||cc|cc}
\hline

\multirow{3}{*}{\textbf{Dataset}}                       & \multicolumn{2}{c||}{\textbf{Proposed}} & \multicolumn{2}{c|}{\multirow{2}{*}{\textbf{Reduction}  \textbf{vs \cite{Mubarik:MICRO:2020:printedml}}}}  & \multicolumn{2}{c}{\multirow{2}{*}{\textbf{Reduction} \textbf{vs \cite{isqed_dt}}}} \\ \hhline{~|--||~~|~~}
                                       &                    &                   & & & & \\ 
\multirow{-2}{*}{\textbf{}} &
  \multirow{-2}{*}{\begin{tabular}[c]{@{}c@{}}\textbf{Area}\tnote{1}\\ ($mm^2$)\end{tabular}} &
  \multirow{-2}{*}{\begin{tabular}[c]{@{}c@{}}\textbf{Power}\tnote{1}\\ ($mW$)\end{tabular}} &
  \textbf{Area}\tnote{1} &
  \multicolumn{1}{c|}{\textbf{Power}\tnote{1}} &
  \textbf{Area}\tnote{1} &
  \textbf{Power}\tnote{1} \\ \hhline{-|--||--|--} 
\textbf{WhiteWine}                     & 11.99              & 1.26              & 21.8x   & \multicolumn{1}{c|}{11.3x}  & 10.5x             & 4.3x             \\
{\color[HTML]{000000} \textbf{Cardio}} & 10.13              & 0.88              & 11.3x   & \multicolumn{1}{c|}{14.1x}  & 4.4x              & 2.4x             \\
\textbf{Arrhythmia}                    & 16.24              & 0.85              & 4.9x    & \multicolumn{1}{c|}{14.1x}  & 1.5x              & 1.3x             \\
\textbf{Balance-Scale}                 & 4.92               & 0.35              & 6.2x    & \multicolumn{1}{c|}{8.2x}   & 5.8x              & 3.6x             \\
\textbf{Vertebral-3C}                  & 2.71               & 0.17              & 6.2x    & \multicolumn{1}{c|}{16.2x}  & 3.4x              & 2.7x             \\
\textbf{Seeds}                         & 3.26               & 0.27              & 8.4x    & \multicolumn{1}{c|}{11.9x}  & 1.2x              & 1.1x             \\
\textbf{Vertebral-2C}                  & 2.22               & 0.15              & 7.4x    & \multicolumn{1}{c|}{18.5x}  & -\tnote{2}                 & -\tnote{2}                \\
\textbf{Pendigits}                     & 89.00              & 6.12              & 3.0x    & \multicolumn{1}{c|}{2.8x}   & 4.2x              & 2.6x             \\ \hhline{-|--||--|--} 

\cellcolor[HTML]{EFEFEF}\textit{\textbf{Average}}  & \cellcolor[HTML]{EFEFEF}17.56              & \cellcolor[HTML]{EFEFEF}1.26              & \cellcolor[HTML]{EFEFEF}8.6x    & \multicolumn{1}{c|}{\cellcolor[HTML]{EFEFEF}12.2x}   & \cellcolor[HTML]{EFEFEF}4.4x              & \cellcolor[HTML]{EFEFEF}2.6x \\ \hline
\end{tabular}
\begin{tablenotes}\scriptsize
\item[] $^1$Total area and total power, including ADCs.\quad $^2$Not evaluated in ~\cite{isqed_dt}.
\vspace{-2ex}
\end{tablenotes}
\end{threeparttable}
\label{tab:comparison}
\end{table}

Finally, Table~\ref{tab:comparison} evaluates the effectiveness of our co-design framework in generating self-powered printed Decision Trees, considering up to $1$\% accuracy loss.
In Table~\ref{tab:comparison}, we also compare our Decision Trees against the baseline exact~\cite{Mubarik:MICRO:2020:printedml} (i.e., Table~\ref{tab:baselines}) and the approximate ones of~\cite{isqed_dt} (with up to $1$\% accuracy loss).
Note that for~\cite{Mubarik:MICRO:2020:printedml} conventional $4$-bit ADCs are used, whereas for~\cite{isqed_dt}, since precision scaling is applied, the smallest suitable conventional ADC for each input is used.
To the best of our knowledge, these are the only works that have investigated printed Decision Trees.
As shown, our Decision Trees achieve on average $8.6$x and $12.2$x lower area and power, respectively, compared to~\cite{Mubarik:MICRO:2020:printedml}.
Similarly, compared to~\cite{isqed_dt}, the corresponding  gains are $4.4$x and $2.6$x, respectively.
Note that in some cases (i.e., BS, V3, and PD), \cite{isqed_dt} features higher area and power consumption compared to our baseline~\cite{Mubarik:MICRO:2020:printedml} due to the use of deeper trees in~\cite{isqed_dt} to compensate for the accuracy loss caused by their applied approximation.
Concluding, as demonstrated by Table~\ref{tab:comparison} all our classifiers except for Pendigits feature power consumption well bellow $2$mW, even when accounting the significant cost of ADCs.
Pendigits also adhered to the $2$mW power constraint but at a $10$\% accuracy loss.
This analysis suggests that our co-design framework can be used to effectively target printed applications with similar complexity to the 
datasets in Table~\ref{tab:comparison}, even after considering the cost of sensors, which is negligible compared to the hardware overheads of printed classifiers.
For instance, relevant sensors reviewed in~\cite{Bleier:ISCA:2020:printedmicro} for such printed applications consume only $5\mu$\!W.
Hence, for the examined datasets, the power increase due to sensors is less than $0.11$mW.
As a result, for less than $1$\% accuracy loss, our framework can efficiently produce printed classifiers, even with $207$ comparators and $11$ inputs, that demand less than $2$mW of power, being thus suitable for printed energy harvester operation~\cite{pintedharvester}.

\section{Conclusion}
Printed electronics offer a solution to extend computing into application domains previously untouchable by silicon-based technologies.
In this work, we propose a co-design framework for printed Decision Trees that incorporates fully parallel unary architectures, bespoke ADCs tailored for such architectures, and an ADC-aware training approach aimed at minimizing the overall hardware cost of on-sensor processing.
Our circuits achieve substantial area and power savings, paving the way towards digital, self-powered, on-sensor printed ML classifiers.


{\small
\section*{Acknowledgment}
This work is partially supported by EU Horizon research and innovation programme, under project CONVOLVE, grant agreement No 101070374, by the European Research Council (ERC), and co-funded by the H.F.R.I call “Basic research Financing (Horizontal support of all Sciences)” under the National Recovery and Resilience Plan “Greece 2.0” (H.F.R.I. Project Number: 17048)
}

\makeatletter
\def\footnoterule{\kern-3\p@
  \hrule \@width 0.75in \kern 2.6\p@} 
\makeatother


{
\bibliographystyle{IEEEtran}
\bibliography{references}
}

\end{document}